\documentclass[letterpaper]{article} 
\usepackage[]{aaai24}  
\usepackage{times}  
\usepackage{helvet}  
\usepackage{courier}  
\usepackage[hyphens]{url}  
\usepackage{graphicx} 
\usepackage{natbib}  
\usepackage{caption} 
\usepackage{algorithm}
\usepackage{algorithmic}

\usepackage{dsfont}
\usepackage{comment}
\usepackage{graphicx}
\usepackage[mathscr]{eucal}
\usepackage{xspace}

\usepackage{amsmath}
\usepackage{amsfonts}
\usepackage{multirow}
\usepackage{graphicx}

\usepackage{mathtools}
\usepackage{cuted}

\usepackage{newfloat}
\usepackage{listings}
\DeclareCaptionStyle{ruled}{labelfont=normalfont,labelsep=colon,strut=off} 
\floatstyle{ruled}
\newfloat{listing}{tb}{lst}{}
\floatname{listing}{Listing}
\title{Heterogeneous Federated Learning via Personalized Generative Networks}
\author{
    Zahra Taghiyarrenani,
    Abdallah Alabdallah, 
    S\l{}awomir Nowaczyk,
    Sepideh Pashami 
}
\affiliations{Center for Applied Intelligence Systems Research, Halmstad University, Halmstad, Sweden \\\{firstname.lastname\}@hh.se} 

\begin{document}

\maketitle

\begin{abstract}
Federated Learning (FL) allows several clients to construct a global machine-learning model without having to share their data. However, FL often faces the challenge of statistical heterogeneity between the client's data. The heterogeneity of the clients will affect the global model's effectiveness. In this paper, utilizing the Domain Adaptation concept, we define an upper bound for the risk of a global model for every single client and theoretically prove that minimizing heterogeneity between clients minimizes this upper bound. We mainly focus on the heterogeneity in the distribution of data across feature spaces of the clients rather than merely considering imbalanced classes, which have mostly been studied in the literature until now. Therefore, with a primary focus on cross-silo settings, we propose a method called FedGenP, which uses server-trained, client-specific generators for knowledge transfer between clients. Each generator aims to provide samples for the corresponding client to minimize its discrepancy with other clients. Experiments conducted on synthetic and real data, along with a theoretical study, support the effectiveness of our method, demonstrating that the generated data helps in constructing a well-generalized global model. In addition, because of the generators, after fine-tuning the global model with both the local and generated data, the clients need not lose the information from others, resulting in well-generalized personalized models as well.
\end{abstract}

\section{Introduction} 
Federated learning (FL) is an approach to train machine learning models while the data are distributed between different clients (data owners). A typical setting in FL has a server organizing the clients to learn a model, namely the global model, collaboratively.  FedAvg is the most well-known FL introduced by \cite{mcmahan2017communication}. In FedAvg, the server receives the local models (models the clients train locally) and calculates their average as a global model. The server sends the global model back to the clients. This communication, namely the federated communication step, will be repeated until the convergence of the global model. 

Considering a federated system, heterogeneity may be present at different levels, including data, model, and communication \cite{ye2023heterogeneous}. For this study, we specifically focus on a system with heterogeneity at the data level. In other words, we address statistical differences in the data that is available among the clients.
The statistical heterogeneity in clients' data affects the performance of the global model; the global model may not uniformly benefit all clients due to this phenomenon. While improving performance for specific clients, others may be disadvantaged by participating in the federated process. By this motivation, minimizing the negative effects of the heterogeneity of the data on the performance of the federated model(s) is a hot topic in the FL community.

We examine the challenge of heterogeneous clients in FL from a Domain Adaptation (DA)  \cite{blitzer2007learning} perspective. The central assumption in machine learning is that training and test data are independently and identically distributed, so-called the IID assumption. However, in reality, data may originate from different domains and violate this fundamental assumption. DA is an approach that reduces the distribution gap between the domains, namely Source and Target \cite{taghiyarrenani2022learning}. In the context of DA, the objective is to build a model specifically for the target domain. However, due to insufficient data from the target domain, the approach involves utilizing available data from the source domain. Given the differences between the domains, DA minimizes the distribution gap between source and target data, ensuring the Independent and Identically Distributed (IID) assumption for training a model.

In the FL context, data originates from different clients, which we equate to the domains in the DA context as in \cite{taghiyarrenani2023analysis}. Therefore, it is probable that they follow different distributions and affect the global model results. In this regard, we first theoretically show that the heterogeneity between one client and the rest of the clients affects the performance of the global model on that client. We construct this discussion from DA's point of view.  However, a key distinction lies in the fact that while DA assumes data from different domains being available, FL does not permit such access to data from the clients.

In dealing with data from diverse clients, we can create a shared model by averaging local models, as in FedAvg \cite{mcmahan2017communication}. However, this approach is significantly impacted by data heterogeneity. On the opposite end of the spectrum is constructing completely personalized models for each client, which, however prevents the exchange of information between them \cite{tan2022towards}. Finding a balance between these extremes, Personalized FL aims to gain knowledge from different clients while constructing personalized models for each client to mitigate bias introduced by vastly different clients \cite{tan2022towards}.

Many different personalized approaches have been proposed in the literature to handle the heterogeneity of the clients, with a recent survey providing comprehensive overview \cite{tan2022towards}. One particularly relevant technique among them is Data Augmentation. Data augmentation is widely used in machine learning as it enhances the diversity of samples, contributing to improved model generalizability. This technique is also beneficial when dealing with data from various domains \cite{taghiyarrenani2022noise}, as it contributes to reducing the disparities between them. 
Data augmentation can also be utilized in the FL setting. However, many existing data augmentation methods mainly focus on addressing class imbalances among clients by augmenting the clients' data to achieve a class balance \cite{tan2022towards}.

This paper proposes data augmentation techniques in the FL setting that minimize the discrepancy between the clients.  Building on theoretical discussions, our objective is to diminish heterogeneity between each client and the rest through data augmentation. Unlike previous approaches, our aim is to address and reduce heterogeneity resulting from data distribution between clients.
To this end, we propose a method (FedGenP) in which the server trains personalized generators for each client based on the local models of other clients. In this way, by generating new synthetic data and augmenting the client's data, the generator will transfer knowledge from other clients. The generators are designed to create data in areas of space where there are discrepancies between clients. However, in order to keeping privacy, they generate data in a latent space instead of the input space. By doing so, and aligning with the theoretical discussions, our goal is to remove heterogenity between the clients and consequently improve the performance of the global model for every client.

\section{Related works}
Heterogeneous FL has been studied in the literature from different perspectives, and different solutions have been proposed as described in ~\cite{gao2022survey, ma2022state}. Authors in \cite{gao2022survey} survey all types of heterogeneity in FL, including statistical heterogeneity, and categorize the approaches to tackling the statistical heterogeneity into two groups: single global and personalized local models. Authors in \cite{ma2022state}, however, just focus on the statistical heterogeneity and categorize different types of heterogeneity and also different approaches to solve the problem. They organize different solutions into four distinct groups: Data-based, Model-based, Algorithm-based, and Framework-based. In addition, authors in \cite{tan2022towards} discuss the personalized FL methods and categorize the solution into 1) global model personalization and 2) Learning personalized models.

Data augmentation is a Data-based approach for personalization to smooth statistical heterogeneity between the clients \cite{ye2023heterogeneous}. In the literature, several data augmentation methods are proposed to solve the heterogeneity of the clients. In some of the existing proposed data augmentation methods, the clients need to share some data \cite{zhao2018federated, jeong2018communication}, or statistical information of the local data \cite{duan2020self} or server needs a proxy data that follows the distribution of the global data. \cite{yoon2021fedmix} proposed a data augmentation based on Mixup \cite{zhang2017mixup}  for FL, where the clients send and receive the averaged data. In addition, the methods consider the heterogeneity because of the imbalanced classes rather than the distribution shift in data samples \cite{zhao2018federated, jeong2018communication, duan2020self, wu2020fedhome}. 

The study in \cite{de2022mitigating} is connected to our work as it links the idea of Domain Generalization to FL. It suggests that using data augmentation helps reduce heterogeneity among clients. However, our approach differs in that they perform data augmentation on each client, such as rotating images, without explicitly addressing the shift between clients.

The methods proposed in \cite{jeong2018communication, wu2020fedhome, wen2022communication, chen2023fraug} augment the data by training a generator and, from this regard, relate to ours. For training generators in a federated setting, the application of knowledge distillation as a framework-based solution holds promising potential.

Authors in \cite{li2023feddkd} propose a decentralized knowledge distillation model to guide the global model by distilling the knowledge of local models.
Authors in \cite{yao2021local} also use knowledge distillation techniques to guide local models by the ensemble of previous global models. Applying knowledge distillation is also suggested by \cite{huang2022learn} to address the catastrophic forgetting problem and to prevent forgetting. To this end, the authors suggest using a public dataset as the proxy dataset. One-shot knowledge distillation is suggested by \cite{gong2022preserving}; It uses an unlabeled public dataset as a proxy dataset. Therefore, one main limitation is using a proxy dataset. A data-free knowledge distillation method is proposed in~\cite{zhu2021data}. The authors consider the problem from a DA perspective and prove the generalization performance of the global model on the global test data. Thus, this method is the most related to our work, despite the fact that they solve imbalance classes rather than data heterogeneity in feature space.

\section{Notations and Preliminaries}
We mainly follow the notations in~\cite{ben2006analysis} and ~\cite{zhu2021data}. We have also summarized all the notations used in the paper in table~\ref{tab: The description of symbols used throughout the paper}.

Let's define $\mathcal{X}$ as input space, $\mathcal{D}(x)$ as the probability distribution on $x \in \mathcal{X}$, and $\hat{\mathcal{D}}(x)$ as an empirical probability distribution. We consider a classification problem with the output space $\mathcal{Y}$ so that we define the ground-truth labeling function $f:\mathcal{X}\to \mathcal{Y}$.

We call triple $\tau = \left\langle \mathcal{D}, \mathcal{X},f \right\rangle$ as a domain; a distribution $\mathcal{D}(x)$ over $x \in \mathcal{X}$ with the corresponding ground-truth labeling function $f$. We aim to construct a classifier model, $\hat{f}_{\theta}$, parameterized by $\theta$. Given $\mathcal{Z}$ as a latent feature space, the classifier $\hat{f}_{\theta}=r_{\theta^r} \circ p_{\theta^p}$ consists of a representation model, $r_{\theta^r}:\mathcal{X}\to \mathcal{Z}$, parameterized by $\theta^r$ and a predictor, $p_{\theta^p}:\mathcal{Z}\to \mathcal{Y}$, parameterized by $\theta^p$; therefore $\theta := [\theta^r;\theta^p]$. Given $l$ as a loss function, we define the risk of the model as $\varepsilon_{\tau}(\theta) =  \mathbb{E}_{x\sim \mathcal{D}(x)}\big[ l\left( \hat{f}_{\theta}(x), f(x) \right) \big]$.

\paragraph{Federated Learning.} We consider a federated setting with one server and $K$ clients, $\mathcal{T} = \left\{ \tau_k \right\}_{k=1}^K$ such that $\tau_k = \left\langle \mathcal{D}_k, \mathcal{X}_k,f_k \right\rangle$ defines the k$^\text{th}$ client domain. This paper assumes all clients have the same input space and labeling function. Accordingly, the definition of $\tau$ will be $\tau_k = \left\langle \mathcal{D}_k, \mathcal{X},f \right\rangle$. We define the risk of every local model as $\varepsilon_{\tau_k}(\theta_k) =  \mathbb{E}_{x\sim \mathcal{D}_k(x)}\big[ l\left( \hat{f}_{\theta_k}(x), f(x) \right) \big]$ where $\hat{f}_{\theta_k}(x)$ is the model trained on the local data of client $\tau_k$. The aim of FL is to learn a global classifier model parameterized by $\theta$. The risk of this model is calculated over all clients $\mathcal{T}$ as $\varepsilon_{\mathcal{T}} (\theta) = \mathbb{E}_{\tau_k \in \mathcal{T}} \left[ \varepsilon_{\tau_k} (\theta) \right]$.



\begin{table*}
  \caption{The description of symbols used throughout the paper}
  \label{tab: The description of symbols used throughout the paper}
  \centering
  \begin{tabular}{lll}              
    Symbol & Formal definition & Description \\ \hline\hline
$\mathcal{X}$ & $\mathcal{X} \subset  \mathbb{R}^p$ & Input space\\ 
\hline
$\mathcal{Y}$ & $\mathcal{Y} \subset  \mathbb{R}$ & Output space\\ 
\hline
$\mathcal{Z}$& $\mathcal{Z} \subset  \mathbb{R}^d$ & Latent space\\
\hline
$\mathcal{D}$ & $\mathcal{D}(x)$ & Probability distribution\\
\hline
$\hat{\mathcal{D}}$ & $\hat{\mathcal{D}}(x)$  & Empirical probability distribution\\
\hline
$\tau$& $\tau := \left\langle \mathcal{D}, \mathcal{X},f \right\rangle$& 
Domain\\
\hline
$r$& $r : \mathcal{X} \to \mathcal{Z}$& Representation function\\
\hline
$p$ & $p:\mathcal{Z}\to \mathcal{Y}$& Predictor function\\
\hline
$\textit{f}$ &  $\textit{f} : \mathcal{X} \to \mathcal{Y}$& Labeling rule\\
\hline
$\acute{\textit{f}}$ & $\acute{\textit{f}} : \mathcal{R} \to \mathcal{Y}$  & Induced image of $\textit{f}$ under $\mathcal{R}$\\\hline
  \end{tabular}
\end{table*}

\section{Theoretical Motivation}

A hypothesis $h$ is defined as $h:\mathcal{X}\to \mathcal{Y}$. $\mathcal{H}$ is a hypothesis space for $\mathcal{X}$. The risk of the hypothesis, $\varepsilon(h)$, captures the disagreement between the hypothesis and the labeling function as $$\epsilon(h) =  \mathbb{E}_{x\sim D}\big[ \left| h(x) - f(x) \right| \big].$$
Similarly, the empirical hypothesis risk is defined as
\begin{center}
   $ \hat{\epsilon}(h) =  \mathbb{E}_{x\sim \hat{\mathcal{D}}}\big[ \left| h(x) - f(x) \right| \big]$.
\end{center}

\paragraph{Symmetric difference hypothesis space $\mathcal{H}\Delta\mathcal{H}$ \cite{blitzer2007learning}:}Given a  hypothesis space $\mathcal{H}$, $\mathcal{H}\Delta\mathcal{H}$ is defined as
\begin{center}
   $ \mathcal{H}\Delta\mathcal{H} = \left\{ h(x) \oplus h'(x) : h,h' \in \mathcal{H} \right\}$,
\end{center}
where $\oplus$ represents the XOR operator. This means a hypothesis belongs to $\mathcal{H}\Delta\mathcal{H}$ if a given pair in $\mathcal{H}$, $h(x)$ and $h'(x)$ disagree.

$\mathcal{A_H}$ is a set of measurable subsets for some hypothesis $h\in \mathcal{H}$, so that,
\begin{center}
   $ \left\{ x: x\in \mathcal{X}, h(x) = 1 \right\} \in \mathcal{A_H}, \forall h\in \mathcal{H} $.
\end{center}
Similarly, $\mathcal{A}_{H \Delta H}$ is defined as 
\begin{center}
   $\left\{ x: x\in \mathcal{X}, h(x) \neq h'(x) \right\} \in \mathcal{A}_{H \Delta H},  \forall h, h' \in \mathcal{H} \Delta \mathcal{H}$.
\end{center}

\paragraph{$\mathcal{H}$-distance between two distributions \cite{blitzer2007learning}:} Given $D$ and $D'$ as two arbitrary distributions, $\mathcal{H}$-distance is defined as:

\begin{center}
   $ d_\mathcal{H} (D,\mathcal{D}') := 2  \sup _{\mathcal{A\in A_H}}\left| \mathrm{Pr}_\mathcal{D} (\mathcal{A}) - \mathrm{Pr}_\mathcal{D'} (\mathcal{A}) \right|$,
\end{center}
and, similarly, $d_{\mathcal{H} \Delta \mathcal{H}} (\mathcal{D},\mathcal{D}')$ is defined as distribution divergence induced by $\mathcal{H} \Delta \mathcal{H}$ as:

\begin{center}
   $ d_{\mathcal{H} \Delta \mathcal{H}} (\mathcal{D},\mathcal{D}') := 2  \sup _{\mathcal{A \in A} _ {\mathcal{H} \Delta \mathcal{H}}}
\left| \mathrm{Pr}_\mathcal{D} (\mathcal{A}) - \mathrm{Pr}_\mathcal{\mathcal{D}'} (\mathcal{A}) \right|$.
\end{center}

The defined distribution divergence is defined over two arbitrary distributions, $D$ and $D'$. Let's define a representation function as $\mathcal{R}: \mathcal{X} \to \mathcal{Z}$ and $\tilde{D} $ and $ \tilde{D'}$ the corresponding distributions over $\mathcal{Z}$. so, one can calculate the distribution divergence over $\mathcal{Z}$ as: 
\begin{center}
   $ d_{\mathcal{H} \Delta \mathcal{H}} (\tilde{\mathcal{D}},\tilde{\mathcal{D}'}) := 2  \sup _{\mathcal{A \in A} _ {\mathcal{H} \Delta \mathcal{H}}}
\left| \mathrm{Pr}_\mathcal{\tilde{\mathcal{D}}} (\mathcal{A}) - \mathrm{Pr}_\mathcal{\tilde{\mathcal{D}'}} (\mathcal{A}) \right|$.
\end{center}

As we aim to discuss the problem from the DA point of view, we will construct our analysis considering the adaptation problem. To this end, we follow~\cite{ben2006analysis, zhu2021data}.

First, we define a domain, $\tau$ as triple $\tau = \left\langle \mathcal{D}, \mathcal{X},f \right\rangle$; a distribution $\mathcal{D}$ over input space $\mathcal{X}$ with the corresponding labeling function $f$. 

\paragraph{Lemma 1. Generalization Bounds for DA \cite{ben2006analysis}:}
Let us assume two different source and target domain as $\tau_s = \left\langle \mathcal{D}_s, \mathcal{X},f \right\rangle$, $\tau_t = \left\langle \mathcal{D}_t, \mathcal{X},f \right\rangle$. Assuming $\mathcal{R}: \mathcal{X} \to \mathcal{Z}$ as a shared representation function for source and target domains, $ \tilde{\mathcal{D}_s}$ and $\tilde{\mathcal{D}_t} $ are corresponding distributions over $\mathcal{Z}$. The generalization bound on the target domain error is defined as follows, so that with the probability at least $1-\delta$, for every $h \in \mathcal{H}$:

\begin{equation}
\label{Eq: Generalization bound for DA}
\begin{split}
    \epsilon_{\tau_t}(h) \le \hat{\epsilon}_{\tau_s}(h) &+ \sqrt{\frac{4}{m}\left(d\log\frac{2em}{d} + \log\frac{4}{\delta} \right)} \\& +d_{\mathcal{H}\Delta \mathcal{H}}(\tilde{\mathcal{D}_s}, \tilde{\mathcal{D}_t}) + \lambda
\end{split}
\end{equation}
where $d$ is the VC-dimension of a set of hypothesis $\mathcal{H}$, $m$ is the number of source samples, $\hat{\epsilon}_{\tau_s}(h)$ is the empirical risk of the hypothesis trained on source data, and $e$ is the base of the natural algorithm. Considering $h^*$ as the optimal hypothesis so that $$h^* = \underset{h\in \mathcal{H}}{\mathrm{argmin}}(\epsilon_{\tau_s}(h) + \epsilon_{\tau_t}(h)),$$$\lambda$ is the optimal risk on two domains, $\lambda = \epsilon_{\tau_s}(h^*) + \epsilon_{\tau_t}(h^*)$.

In the FL setting, we consider every client as a domain. Therefore, we define the set of clients as $\mathcal{T} = \left\{ \tau_k: k\in K \right\}$ with $K$ clients where $\tau_k = \left\langle \mathcal{D}_k, \mathcal{X}, f \right\rangle$. Accordingly, all clients share the same input space and the same labeling function (all local data belong to one global distribution). However, the distribution of the client's data is different.

We perform analysis for a random client, $\left\{\tau_{k'}:k'\in K\right\}$. The global hypothesis is the ensemble of the user predictors, i.e., $h_g = \frac{1}{K}(\sum_{k=1}^{K}h_k)$, where $h_k$ means the hypothesis learned on domain $\tau_k$. 

According to Jensen inequality,
\begin{equation}
\begin{split}
\label{Eq: Jensen 0}
 \epsilon_{k'}(h_g) &=   \epsilon_{k'} (\frac{1}{K}\sum_{k=1}^{K} h_k) 
 \le  \frac{1}{K}  \sum_{k=1}^{K} \epsilon_{k'}(h_k),\\
\end{split}
\end{equation}
the generalization performances of all $h_k$ on the $k'$ affect the performance of the global model $h_g$ on domain $k'$.

Since we want to understand the effect of users on each other, for the rest of the analysis, we consider client $k'$ as the target domain and any other client as the source domain.  
For any paired clients $k$ and $k'$, according to Equation~\ref{Eq: Generalization bound for DA}, one can conclude that with the probability of at least $1-\delta_k$,
\begin{equation}
\label{Eq: Generalization bound for paired clients 0}
\begin{split}
    \epsilon_{\tau_{k'}}(h_k) \le \hat{\epsilon}_{\tau_k}(h_k) +& \sqrt{\frac{4}{m}\left(d\log\frac{2em}{d} + \log\frac{4}{\delta_k} \right)} \\&+ d_{\mathcal{H}\Delta \mathcal{H}}(\tilde{\mathcal{D}}_k, \tilde{\mathcal{D}}_{k'}) + \lambda_k,
\end{split}
\end{equation}
where $k\in K$ and $\lambda_k$ means the optimal risk on clients ${k'}$ and $k$. Equation~\ref{Eq: Generalization bound for paired clients 0} holds true for all clients $k\in K$ with probability at least $1-\delta_k$. For the rest, we replace $\delta_k$s with $\delta$ where $\delta = \min(\delta_1, ..., \delta_K)$. 
Furthermore, we replace $\delta$ with $\frac{\delta}{K}$. Accordingly, we conclude that for any paired clients $k$ and $k'$  with the probability of at least $1-\frac{\delta}{K}$,

 \begin{equation}
\label{Eq: Generalization bound for paired clients 00}
\begin{split}
    \epsilon_{\tau_{k'}}(h_k) \le \hat{\epsilon}_{\tau_k}(h_k) &+ \sqrt{\frac{4}{m}\left(d\log\frac{2em}{d} + \log\frac{4K}{\delta} \right)} \\&+ d_{\mathcal{H}\Delta \mathcal{H}}(\tilde{\mathcal{D}}_k, \tilde{\mathcal{D}}_{k'}) + \lambda_k.
\end{split}
\end{equation}

\paragraph{Theorem 1. Generalization Bounds for clients in FL:}
Let's assume $\tau_{k'} = \left\langle \mathcal{D}_{k'}, \mathcal{X},f \right\rangle$ as an arbitrary client and $\tau_k = \left\langle \mathcal{D}_k, \mathcal{X},f \right\rangle$ as any other client. Assuming $\mathcal{R}: \mathcal{X} \to \mathcal{Z}$ as a shared representation function for all clients, $ \tilde{D_k}$ and $\tilde{D_{k'}} $ are corresponding distributions over $\mathcal{Z}$. Then, with the probability of at least $1-\delta$:

 \begin{equation}
 \label{Eq: theorem Generalization for clients}
 \begin{split}
\epsilon_{\tau_{k'}}(h_g)
\le
\frac{1}{K}\sum_{k\in \left[ \mathcal{T}\right] } \left(\hat{\epsilon}_{\tau_k}(h_k) + d_{\mathcal{H}\Delta \mathcal{H}}(\tilde{\mathcal{D}}_k, \tilde{\mathcal{D}}_{k'}) + \lambda_k \right)\\+ \sqrt{\frac{4}{m}\left(d\log\frac{2em}{d} + \log\frac{4K}{\delta} \right)},
\end{split}
\end{equation}
where  $d$ is the VC-dimension of a set of hypothesis $\mathcal{H}$, $m$ is the number of samples of the client $k$, $\hat{\epsilon}_{\tau_k}(h_k)$ is the empirical risk of the hypothesis trained on data of the client $k$ calculated over client $k$, and $e$ is the base of the natural algorithm. Considering $h^*$ as the optimal hypothesis so that $$h^* = \underset{h\in \mathcal{H}}{\mathrm{argmin}}(\epsilon_{\tau_k}(h) + \epsilon_{\tau_{k'}}(h)),$$ $\lambda$ is the optimal risk on two clients, $\lambda = \epsilon_{\tau_k}(h^*) + \epsilon_{\tau_{k'}}(h^*)$.

\noindent\textit{Proof.}

According to equations~\ref{Eq: Jensen 0} and ~\ref{Eq: Generalization bound for paired clients 00}, and given
\begin{eqnarray}
    &\eta& \equiv \sqrt{\frac{4}{m}\left(d\log\frac{2em}{d} + \log\frac{4K}{\delta} \right)} \\
    &\omega& \equiv \frac{1}{K}\sum_{k\in \left[ \mathcal{T}\right] } \left(\hat{\epsilon}_{\tau_k}(h_k) + d_{\mathcal{H}\Delta \mathcal{H}}(\tilde{\mathcal{D}}_k, \tilde{\mathcal{D}}_{k'}) + \lambda_k \right),
\end{eqnarray}
we conclude
\begin{eqnarray}
    &\mathrm{Pr}& \left( \epsilon_{\tau_{k'}}(h_g) > \omega + \eta \right) \nonumber\\
    &\le&  \mathrm{Pr} \left( \frac{1}{K}  \sum_{k=1}^{K} \epsilon_{\tau_{k'}}(h_k) > \omega + \eta  \right) \nonumber \\
    &\le&  \mathrm{Pr} \left( \bigvee_{k\in \mathcal{T}}^{} \epsilon_{\tau_{k'}}(h_k) 
    > \hat{\epsilon}_{\tau_k}(h_k) + d_{\mathcal{H}\Delta \mathcal{H}}(\tilde{\mathcal{D}}_k, \tilde{\mathcal{D}}_{k'}) + \lambda_k + \eta \right) \nonumber \\
    &\le& K * (\frac{\delta}{K}) = \delta \hspace{5cm}\square
\end{eqnarray}


Therefore, according to Theorem 1, we defined an upper bound for the risk of the $h_g$ on the user $k'$, which depends on the distribution discrepancy of this user's data from the data of the rest of the users.

\section{Our proposed method}
\begin{figure}[t]
\includegraphics[width=8cm]{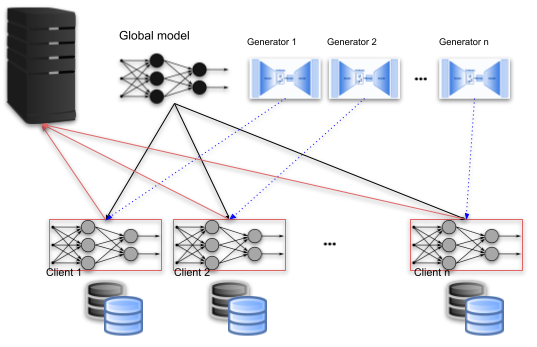}
\caption{Overview of the FedGenP method: Server constructs a global model together with client-specific generators and sends them to the clients. The clients (re)train their local models using both the original and generated data, and then send the trained local model to the server.}
\label{fig: method}
\end{figure}

\begin{figure}[t]
\includegraphics[width=8cm]{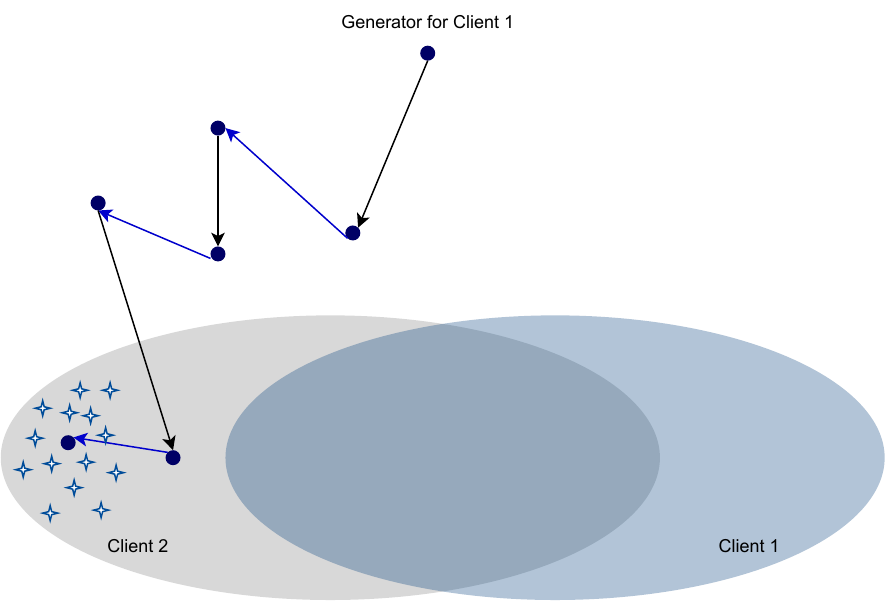}
\caption{Illustration of the intuition behind generator's training process and data generation.}
\label{fig: grl}
\end{figure}

The core idea of this work is minimizing the discrepancy between clients, which adversely affects the performance of the global model. Our method addresses this issue by learning client-specific (personalized) generative models, as demonstrated in Figure \ref{fig: method}. These personalized generative models are constructed using the local models and will generate data in a latent space, which ensures FL's privacy concerns. 
To establish intuition, we present a federated system with two clients in Figure \ref{fig: grl}, representing the training process for a generator associated with Client 1. The black dot points indicate where the generator generates data. The black arrows represent the direction toward the center point of Client 2, while the blue arrows indicate the reverse direction toward the center point of Client 1. First, the generator is one step trained to reach the center point of client 2. Subsequently, in the next step, the generator is trained to reach the center point of Client 1, but during training, we reverse the direction of its gradient; in fact, it will move away from the center point for Client 1. In other words, our approach guides the generator to generate data from areas that Client 1 lacks but which can be provided by Client 2.

The complete process of the proposed FL method (FedGenP) is as follows. At initialization, the algorithm starts at the server, initializing the global model's parameters and broadcasting it to the clients. The clients will train their own models using the data available locally and send the local models back to the server. After the initialization stage, the algorithm iterates until convergence, alternating between server-side and client-side stages.

\paragraph{Server-Side Stage:} At this stage, the server aggregates clients' models into the global model $\theta=\mathds{E}_{k \in \{1..K\}}(\theta_k)$. Moreover, the server uses the clients' local models' parameters $\theta^p_k$ to train client-specific generators $\{G_k(z|y;\theta^g_k)\}^K_{k=1}$. A generator $k$ is trained to generate examples in the regions of space where heterogeneity occurs between the client $k$ decision $P(y|z;\theta^p_k)$ and the decision of the other clients $\{P(y|z;\theta^p_i):i\neq k\}^K_{i=1}$. Therefore, we aim to find the set of parameters $\theta_k^g$ of the generator for client k so that it maximizes $\{P(y|z;\theta^p_i):i\neq k\}^K_{i=1}$ according to Equation \ref{Eq: Generator_max}. This means the generator generates data in $z$ in such a way that this generated data is classified as the desired class by all clients \textit{except} client k.
In addition, we aim to find the set of parameters $\theta_k^g$ of the generator for client k, so that minimizes $P(y|z;\theta^p_k)$ according to equation \ref{Eq: Generator_min}. This means the generator aims for the data that is \textit{not} classified as the desired class by the client k. Therefore, in this way, by two criteria captured by Equations \ref{Eq: Generator_max} and \ref{Eq: Generator_min}, we specifically aim to find conflicts between the local models. Since those conflicts directly represent the data heterogeneity, we ensure that the generated data aligns different clients. 

\begin{equation}
\label{Eq: Generator_max}
\begin{split}
\theta^g_k = & \operatorname*{argmax}_{\theta^g_k} \mathbb{E}_{y\sim\hat{p}(y)}\mathbb{E}_{z\sim G(z|y;\theta^g_k))}\Big[\sum_{i=1}^{K} \mathds{1}(i \neq k)\log p(y|z; \theta_i^p)\Big] 
\end{split}
\end{equation}

\begin{equation}
\label{Eq: Generator_min}
\begin{split}
\theta^g_k = & \operatorname*{argmin}_{\theta^g_k} \mathbb{E}_{y\sim\hat{p}(y)}\mathbb{E}_{z\sim G(z|y;\theta^g_k))}\Big[\log p(y|z; \theta_k^p)\Big] 
\end{split}
\end{equation}

It is notable that we consider a uniform distribution for $\hat{p}(y)$. Hence, the server only needs to know the number of classes, not their true distributions. This is particularly important since we aim to minimize the need for clients to share information.
For learning the generator $G_k$ with parameters $\theta^g_k$, for every client $i$, we define 

\begin{equation}
\label{Eq: loss_client_i}
\begin{split}
J^i(\theta^g_k) = \mathbb{E}_{y\sim\hat{p}(y)}\mathbb{E}_{z\sim G(z|y;\theta^g_k))} l^i(p_i(z),y),
\end{split}
\end{equation}
where $p_i$ is the predictor function of the local model of the client $i$, and $l$ calculates the disagreement between the output of the predictor function of the local models and the desired class, $y$, for which the generator will generate data. 
Using equation \ref{Eq: loss_client_i}, we define the $J_1(\theta^g_k)$ and $J_2(\theta^g_k)$ objectives as:

\begin{equation}
\label{Eq: loss_generator1}
\begin{split}
J_1(\theta^g_k) = \sum_{i=1}^{K} \mathds{1}(i \neq k) J^i(\theta^g_k),
\end{split}
\end{equation}

\begin{equation}
\label{Eq: loss_generator2}
\begin{split}
J_2(\theta^g_k) =  J^k(\theta^g_k).
\end{split}
\end{equation}

For training the generator using stochastic gradient descent, we need to minimize  $J_1(\theta^g_k)$ while maximizing $J_2(\theta^g_k)$. For maximizing $J_2(\theta^g_k)$, we use Gradient Reversal Layer (GRL) \cite{ganin2016domain}. In training, $J_1(\theta^g_k)$  will be minimized in one round. In the next round, we use the GRL layer and reverse the gradient direction while minimizing $J_2(\theta^g_k)$. This process is illustrated in Figure \ref{fig: grl}.

Furthermore, we adopt a Variational Autoencoder (VAE) approach for the generator in the latent space. Instead of directly generating samples, we generate a distribution and then sample data from this learned distribution. After constructing generators, the server will send them to the corresponding clients. 

\paragraph{Client-Side Stage:} Each client $k$ generates $m_k$ examples $Z^{g}_k$ in the $\mathcal{Z}$ space, sampling from the generator $G_k(z|y;\theta^g_k)$. The client uses the generated examples $Z^g_k$ along with its local examples $X_k$ to retrain the local model. Therefore, the object function of the local client $k$ will be designed to minimize the risk of the model on both real and generated data. In other words, the model will be trained to maximize $P(y|z^{g}_k;\theta^{p}_k)+P(y|z_k;\theta^{p}_k)P(z_k|x_k;\theta^{f}_k)$ for a certain number of iterations $T$. The generated examples will help a client's local model gain information about the decision boundaries of other clients' models, compensating for its limited scope of the global space $\mathcal{X}$. The client sends the local model's parameters $\theta_k=[\theta^f_k, \theta^p_k]$ to the server.

Details of FedGenP are summarized in Algorithm~\ref{alg: Algorithm}.

\begin{algorithm}[t]
\caption{\textsc{FedGenP}}
\label{alg: Algorithm}
\begin{algorithmic}[1] 
\REQUIRE Server Side: Clients $\mathcal{T}=\{\mathcal{T}_{k}\}^{K}_{k=1}$
\REQUIRE Clients' Side: $\{X_k \subset \mathcal{X}, y_k \in \mathcal{Y}\}^K_{k=1}$; Local steps $\mathrm{T}$
\STATE \textbf{Initialize:} 
\STATE Server initializes Global parameters $\theta$;
\STATE Server broadcast $\theta$ to clients $\mathcal{T}$
\FOR{$\tau \in \mathcal{T}$ in parallel}
\STATE $[\theta^{f}_k, \theta^{p}_k] \gets \text{argmax}_{[\theta^{f}_k, \theta^{p}_k]}(P(y|z_k;\theta^{p}_k)P(z_k|x_k;\theta^{f}_k)), T$)
\STATE $\theta_k \gets [\theta^{f}_k, \theta^{p}_k]$
\STATE Send $\theta_k$ to Server 
\ENDFOR
\REPEAT
    \STATE \textbf{Server-Side:}
    \STATE $\theta \gets \mathds{E}_{k \in \{1..K\}}(\theta_k)$
    \FOR{$k \in \{1..K\}$}
    \STATE $\theta^g_k \gets \text{Opt Equations~\ref{Eq: Generator_max} and \ref{Eq: Generator_min} using } \{\theta^p_k\}^K_{k=1} $
    \ENDFOR
    \STATE Update $G = \{G_{k}(z|y; \theta^g_k): k \in \{1..K\} \}$
    \STATE Server broadcast $\theta$ and $G_k$ to $\tau_k \in \mathcal{T}$ for $k \in \{1..K\}$, 
    
    \STATE \textbf{Client-Side:}
    \FOR{$\tau \in \mathcal{T}$ in parallel}
    \STATE $z^{g}_k \sim G_{k}(z|y; \theta^g_k)$
    \STATE $[\theta^{f}_k, \theta^{p}_k] \gets \text{argmax}_{[\theta^{f}_k, \theta^{p}_k]}(P(y|z^{g}_k;\theta^{p}_k)+P(y|z_k;\theta^{p}_k)P(z_k|x_k;\theta^{f}_k), T$)
    \STATE $\theta_k \gets [\theta^{f}_k, \theta^{p}_k]$
    \STATE Send $\theta_k$ to Server
    \ENDFOR
\UNTIL{Convergence}
\RETURN $\theta$ 

\end{algorithmic}
\end{algorithm}

\section{Experiments}
In this section, we evaluate the performance of our proposed method and compare it with  the following state-of-the-art methods:
\begin{itemize}
    \item \textsc{FedAvg} \cite{mcmahan2017communication} calculates the global model by averaging the local models. 
    \item \textsc{FedGen} \cite{zhu2021data} is the work closest related to ours. The server constructs a shared generator and sends it to clients to remove the distribution distance between the global data and clients' data. 
    \item  \textsc{FedEnsemble} \cite{lin2020ensemble} ensembles the predictions of the local models. We compare our results with this method, as this is a competitive method to \textsc{FedGen}.
    
\end{itemize}

In our experimental setup, we follow Leaf FL benchmark \cite{caldas2018leaf} and use the CelebA dataset, a collection of celebrity faces. The task selected for prediction is a binary classification to predict whether or not the celebrity in the image is smiling. 
The clients are simulated to include different individuals within the dataset. This arrangement inherently introduces discrepancies between the samples available to each client, reflecting the real-world heterogeneity in distributed environments. More specifically, this dataset simulates the heterogeneity for which our method is designed, namely heterogeneity resulting from statistical heterogeneity in the features, not from the imbalanced classes across the clients.

We also follow \cite{zhu2021data} and use MNIST \cite{lecun2010mnist}, which is a digit classification dataset, and EMNIST \cite{cohen2017emnist}, which is a digit and character classification dataset for evaluation. For these datasets, when distributing the dataset between the clients, the degree of heterogeneity (imbalance-classes) is controlled by parameter $\alpha$ of the Dirichlet distribution as in \cite{hsu2019measuring};  A smaller value of $\alpha$ corresponds to higher levels of heterogeneity between the clients. It is important to note that this method of creating heterogeneity results in imbalanced class distributions among different clients. Nevertheless, we also designed experiments in which the amount of data owned by each client is limited by reducing the proportion of the entire training dataset that is distributed among the clients. This is particularly designed because this limited data can also introduce heterogeneity between the client's model.
We report two sets of results for each of the three datasets. The first set of results is the results of the global model on a global test dataset. These results are indicated by the name of the method. For this purpose, the global model is evaluated immediately after aggregation on the global test dataset. 

The second set of results, indicated with ``(Per)'', are personalized results. After retraining the global model using the local (and generated data in our method), every client evaluates it on its own test data. The reported results are the average of the results of all clients.

The network architectures and the used hyper-parameters are summarized in Table \ref{tab: Hyperparameters}.

\begin{table}[]
\centering
\resizebox{\columnwidth}{!}{%
\begin{tabular}{|l|l|l|l|}
\hline
\rotatebox{45}{Dataset} &   \rotatebox{45}{Classification Network }& \rotatebox{45}{Generators Network }& \rotatebox{45}{Other hyperparameters }  \\ \hline

CelebA   &        \begin{tabular}[c]{@{}l@{}} Conv2d(16), MaxPool2d()\\ Conv2d(32), MaxPool2d()\\ Conv2d(64), MaxPool2d()\\ Flatten\\ fc(784), ReLU\\ fc(32), ReLU\end{tabular}                &          \begin{tabular}[c]{@{}l@{}}fc(128), ReLU\\ fc(32), ReLU\end{tabular}         & \multirow{3}{*}{\begin{tabular}[]{@{}l@{}}Learning rate = 0.01\\ Generator Learning rate = 0.01\\ Optimizer = SGD\\ Batch Size = 32\\  Local Epoch = 20 \\ Number of total and active users\\ are specified per experiment.\end{tabular} }     \\

\cline{1-3}
Mnist   &       \begin{tabular}[c]{@{}l@{}}Conv2d(6)\\ Conv2d(16)\\ Flatten\\ fc(784), ReLU\\ fc(32), ReLU\end{tabular}                 &        \begin{tabular}[c]{@{}l@{}}fc(256), ReLU\\ fc(32), ReLU\end{tabular}           &                       \\ \cline{1-3}

EMnist  &        \begin{tabular}[c]{@{}l@{}}Conv2d(6)\\ Conv2d(16)\\ Flatten\\ fc(784), ReLU\\ fc(32), ReLU\end{tabular}                    &        \begin{tabular}[c]{@{}l@{}}fc(256), ReLU\\ fc(32), ReLU\end{tabular}           &                       \\ \hline
\end{tabular}%
}
\caption{Default hyperparameters in the experiments.}
\label{tab: Hyperparameters}
\end{table}

Table \ref{tab: results on CelebA} compares the performance of FedGenP with that of the existing methods on the CelebA dataset. We examine five different setups based on the number of active users versus the total number of clients (r) for this dataset. In all experiments, the total number of clients is 25. 

As can be seen, FedGenP in a personalized version outperformed the other methods significantly, as expected from this dataset that contains a significant amount of heterogeneity in features. The exception pertains to the experiment conducted with the parameter (r) set to 0.2. As the proposed generator in our method works based on all clients' models, when there are not enough active clients in the system, the generator struggles to create samples with adequate diversity. This could explain why, in this particular scenario, the results do not exhibit as significant a margin as in the other cases. 
In a non-personalized analysis, our method's outcomes are also effective compared to FedGen. It's crucial to note that FedGen requires clients to share class statistics $P(y)$, while our method operates without this, increasing privacy. This distinction becomes especially important in cross-silo scenarios, such as involving various companies as clients. In such cases, having access to class-specific statistics, like those related to faulty equipment data, for example, becomes particularly restrictive.

If we compare the results of the FedGen and (Per)FedGen, we can see that there is not a remarkable difference between them. This is because the generator of FedGen generates data in the space where all clients agree. This data effectively aligns the sample space of every client. Thus, while the generator of FedGenP generates data in the part of space where heterogeneity exists, these diversifying samples actually result in local models that agree and are more generalizable. Consequently, it shows better performance even on the cross-client test data.

One thing to note is the weak result of the (Per)FedAvg. According to the reported deviation, which is more than 2, we can conclude that FedAvg cannot converge with only 100 communication steps. 
 
\begin{table*}[]
\centering
\resizebox{0.8\textwidth}{!}{%
\begin{tabular}{|l|lllll|}\hline
 \multicolumn{6}{c|}{CelebA Dataset} \\ \hline \hline
 \% Active users &
  r = 0.2 &
  r = 0.4 &
  r = 0.6 &
  r = 0.8 &
  r = 1 
\\\hline
FedAvg            &  85.44 (1.38)           &   85.07 (1.89)           &  84.84 (1.60)             &  84.80 (1.71)           & 84.65 (1.70)  \\
FedEnsemble       &  86.03 (0.99)           &  86.07 (0.53)            &  86.18 (0.53)             &  86.60 (1.28)           &  85.50 (1.04)       \\
FedGen            &  86.26 (1.23)           &  87.77 (0.85)            &  \textbf{87.38}(1.37)     &   87.29(1.02)           &  \textbf{87.77 (0.89)}  \\
FedGenP (ours)    & \textbf{86.96}(1.03)    &\textbf{87.89}(1.05)      &  87.21 (1.51)             &  \textbf{87.90}(1.21)   &  87.56 (1.23) \\\hline
(Per)FedAvg        & \textbf{90.46 (2.20)}  & 89.84 (1.62)             & 88.84 (0.59)              & 88.49 (0.57)            & 87.65 (0.91)  \\
(Per)FedEnsemble   & 85. 74 (0.57)          &   86.12 (0.73)           & 86. 08 (0.92)             & 86.49 (1.08)            & 86.51 (0.93)  \\
(Per)FedGen        &  86.69 (0.80)           & 88.35 (0.51)             & 87.58 (1.21)              & 87.44 (0.81)            &   88.35 (0.51)   \\
(Per)FedGenP (ours) & 86.98 (0.92)  & \textbf{90.06 (0.25)}    & \textbf{91.72 (0.53)}     & \textbf{90.39 (0.37)}   & \textbf{90.06 (0.25)} \\\hline 
\end{tabular}%
}
\caption{The results on the CelebA dataset. The results provided show the accuracy of the global model after 100 federated communications. Results averaged from 3 experimental iterations. }
\label{tab: results on CelebA}
\end{table*}

\begin{figure}[t]
\includegraphics[width=8cm]{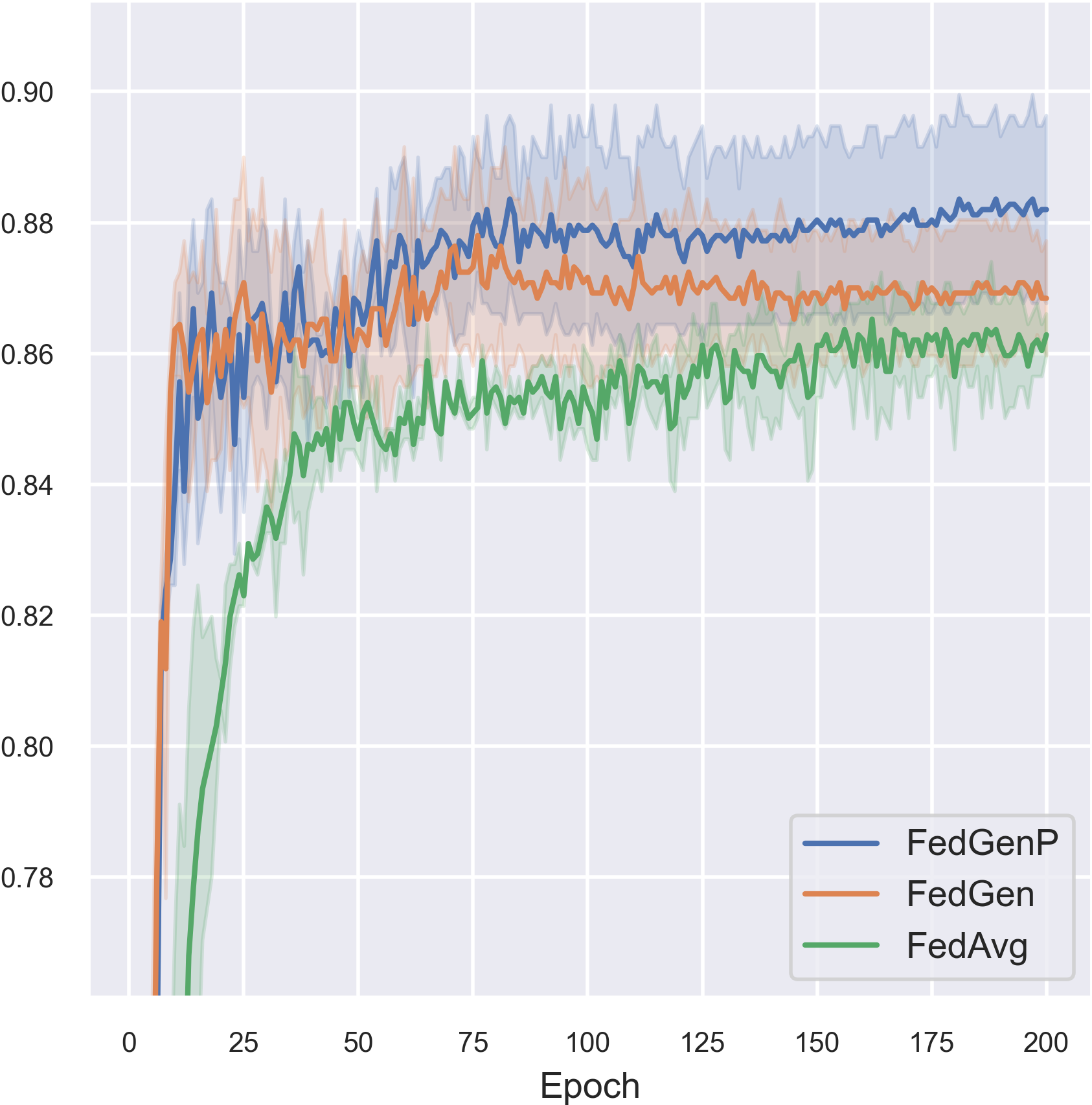}
\caption{The progress of convergence of the global model on the CelebA dataset with r=1.}
\label{fig: CelebA1}
\end{figure}

Among the settings where our method outperforms the rest, we randomly select one and show the progress of the convergence of the global model with 200 federated communications in Figure~\ref{fig: CelebA1}. The convergence of the global model has the same behavior in all cases where it outperforms the rest. Due to this, only one of them is shown here. Figure \ref{fig: CelebA1} illustrates that the proposed method converges faster than FedAvg; however, it is not faster than FedGen. Yet, because FedGenP captures conflicts and resolves them, it improves the final results.

Table \ref{tab: results on Mnist} shows the results on the MNIST dataset. For this dataset, we consider six different settings based on the percentage of the data used as training data, which is distributed among the clients, and the parameter alpha. Alpha controls the heterogeneity between clients by allowing imbalanced classes to be created. Additionally, when clients have less training data distributed among them, the local models may differ and be in conflict with each other, i.e., heterogeneous clients. Using both settings, we explore how our proposed method behaves when heterogeneity is present in different forms.

Our method outperforms other approaches when working with smaller data, such as utilizing only five or ten percent of the training data. This advantage arises from the fact that reduced data can magnify distribution discrepancies between clients, aligning with our method's design focus and thus showcasing its superiority. Nevertheless, in scenarios where each client possesses a substantial amount of data (when utilizing all available training data), the FedGen proves capable of resolving the issue. This is due to the similarity among clients in this situation, resulting in the absence of conflicts. Consequently, our designed generators fail to provide additional benefits to individual clients. Thus, the deployment of one generator, presented by the FedGen, remains sufficient. However, we should note that when compared with FedGen, our method does not require statistics about classes, i.e., $P(y)$. In other words, our method without accessing to such information is competitive to FedGen in the presence of imbalanced classes. 

From table \ref{tab: results on Emnist}, we can see the same behavior on the EMNIST dataset. Our method does not outperform other methods when using all available training data to simulate the clients since there will be no heterogeneity between the samples. However, when there is heterogeneity, and specifically when there is insufficient data to distribute between the clients, our method is superior to the others by removing the heterogeneity. More particularly, our method is efficient on this dataset when utilizing just 10 percent of the training data and setting alpha to 0.01, indicating high heterogeneity.

\begin{table*}[t]
\centering
\resizebox{\textwidth}{!}{%
\begin{tabular}{|l|ll|ll|ll|}\hline
\multicolumn{7}{c}{MNIST Dataset} \\ \hline 
\% Training data - \#Active users  & \multicolumn{2}{c|}{100\% - 20} & \multicolumn{2}{c|}{10\% - 10} & \multicolumn{2}{c|}{5\% - 20} \\ \hline \hline
 Alpha  (Dirichlet distribution) &
  $\alpha$ = 0.1 &
  $\alpha$ = 1 &
  $\alpha$ = 1 &
  $\alpha$ = 10 &

  $\alpha$ = 1 &
  $\alpha$ = 10 \\\hline
FedAvg            &  90.13 (1.92)   &  94.78 (0.51)   &  93.75 (0.62)  &   94.41 (0.43)            &    91.89 (0.71)    &      92.35 (0.45)          \\
FedEnsemble       &  90.86 (0.68)   &  94.13 (0.54)    &  92.56 (0.58)  &   93.77 (0.32)      &           90.81 (0.64)    &        91.13 (0.49)       \\
FedGen            &  \textbf{96.18} (0. 20)   &  \textbf{97.74} (0.06)    &  94.56 (0.15)  &              95.04 (0.06)    & 91.95 (0.51) &      92.38 (0.34)       \\
FedGenP (ours)    & 95.41 (0.40)   &  97.72 (0.06)       & \textbf{95.08} (0.19)     &    \textbf{95.47} (0.04)   & \textbf{92.29 (0.37)} &    \textbf{92.97} (0.31)        \\\hline \hline 
(Per)FedAvg           &90.11 (2.19)&94.83 (0.47)& 94.85 (0.77)&95.44 (0.51)&91.92 (0.70)& 92.35 (0.49)    \\
(Per)FedEnsemble      &   90.84 (0.45)          &    94.19 (0.53)        & 94.24 (0.22) & 94.83 (0.19)& 90.96 (0.15)&   91.74 (0.32)  \\

(Per)FedGen            &    \textbf{96.22 (0.20)} &\textbf{97.79 (0.7)}     &  95.66 (0.17)      &    96.11 (0.20)    & 91.95 (0.51) &       92.38 (0.36)     \\
(Per)FedGenP (ours)    &  95.48(0.40)   &   97.70 (0.09)  &  \textbf{96.13 (0.24)}  &    \textbf{96.54 (0.19)}    &    \textbf{92.30 (0.37) }   & \textbf{92.97} (0.42) \\\hline 
\end{tabular}%
}
\caption{The results on the MNIST dataset. The results are the accuracy of the global model after 300 communication steps. }
\label{tab: results on Mnist}
\end{table*}

\begin{table*}[h]
\centering
\resizebox{0.8\textwidth}{!}{%
\begin{tabular}{|l|ll|ll|}\hline
\multicolumn{5}{c}{EMNIST Dataset} \\ \hline 
\% of Training data  & \multicolumn{2}{c|}{100\%} & \multicolumn{2}{c|}{10\%}  \\ \hline \hline
 Alpha (Dirichlet distribution) &
  $\alpha$ = 0.1 &
  $\alpha$ = 1 &

  $\alpha$ = 0.01 &
  $\alpha$ = 1  \\\hline
FedAvg            &   72.42 (0.62)           &  80.00 (0.48)            &   60.64 (0.59)               &    79.35 (0.34)                     \\
FedEnsemble       &   72.54 (0.73)           &  80.07 (0.45)            &   61.53 (0.39)               &    79.38 (0.34)              \\
FedGen            &  \textbf{76.10} (0.85)   &  \textbf{84.37} (0.63)   &   69.51 (0.16)               &    83.22 (0.14)                   \\
FedGenP (ours)    &   74.82 (0.83)           &  83.64 (0.13)            &   \textbf{70.09} (0.24)      &  \textbf{83.69} (0.18)            \\\hline
(Per)FedAvg       &   68.29 (1.7)            &  79.97 (0.42)            &   68.89 (0.18)                &   79.42 (0.35)              \\
(Per)FedEnsemble  &   66.62 (0.18)           &  79.40 (0.13)            &   68.79 (0.10)                &   79.04 (0.16)                \\
(Per)FedGen       &   \textbf{75.36 (0.79) } &  \textbf{83.89 (0.07) }  &   75.81 (0.12)                &   83.27 (0.13)                \\
(Per)FedGenP (ours)&  75.21 (0.69)           &  83.43 (0.16)            &   \textbf{77.05 (0.12)   }    &   \textbf{83.85 (0.24) }       \\\hline 
\end{tabular}%
}
\caption{The results on the EMNIST dataset. The results are the accuracy of the global model after 100 communication steps.}
\label{tab: results on Emnist}
\end{table*}


Up to this point, we have always used a local epoch of 20 in all of our experiments. However, since our method utilizes generators constructed based on the local model, the convergence of the local models when every client trains it locally significantly impacts the final results. One of the factors contributing to the convergence of local models is the number of local epochs, so we examine how it affects the final results. Therefore, we use the MNIST dataset with 5 percent of the training data distributed to twenty clients, set the alpha to 10, and calculate the output with local epochs 20, 40, 60, 80, and 100. Table \ref{tab: local_epochs} summarizes the results. Clearly, having a higher number of local epochs increases performance. Nevertheless, with fewer communication steps, we achieve nearly equivalent results, highlighting the superiority of our method in the presence of heterogeneity. This holds particular significance in the context of heterogeneous FL. As claimed by the authors in \cite, FedAvg is more likely to address the challenge of heterogeneity with fewer local epochs, although it results in a higher communication cost. In contrast, our method effectively addresses heterogeneity, enhancing results even with a high number of local epochs, ultimately reducing the required communication overhead.

\begin{table}[t]
\centering
\resizebox{0.45\textwidth}{!}{%
\begin{tabular}{|l|llll|}\hline
 \multicolumn{5}{c|}{Mnist Dataset} \\ \hline \hline
 Local epochs &
  20 &
  30 &
  40 &
  50 
\\\hline
Accuracy of global model            &  92.82   &  92.84 &92.95  &  92.99   
\\\hline 
\end{tabular}%
}
\caption{ The effect of the number of Local Epochs on the results of FedGenP. The results are the accuracy of the global model after 50 communication steps.}
\label{tab: local_epochs}
\end{table}

\begin{figure}[h]
    \centering
    \includegraphics[width=2.7cm]{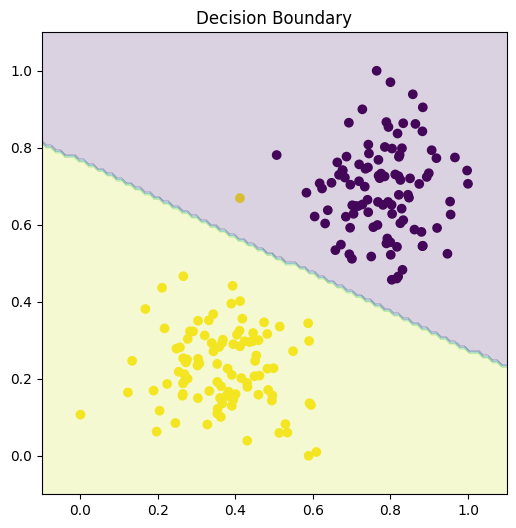}
    \includegraphics[width=2.7cm]{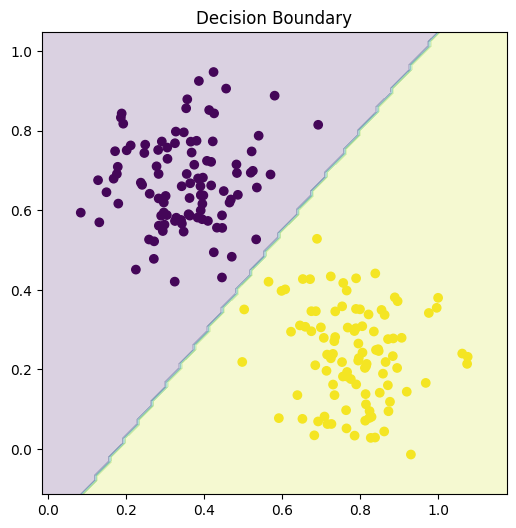}
    \includegraphics[width=2.8cm]{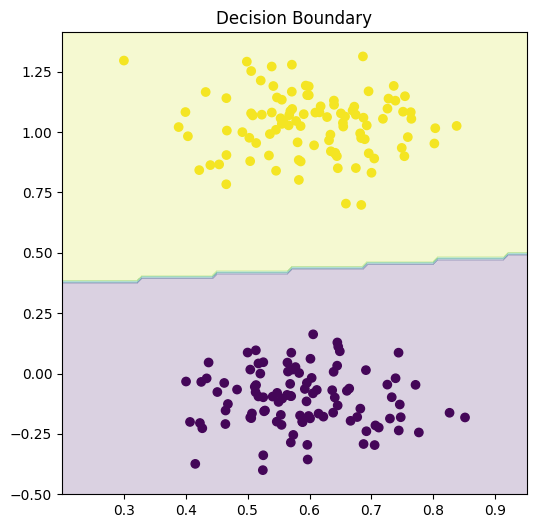}
    \includegraphics[width=2.7cm]{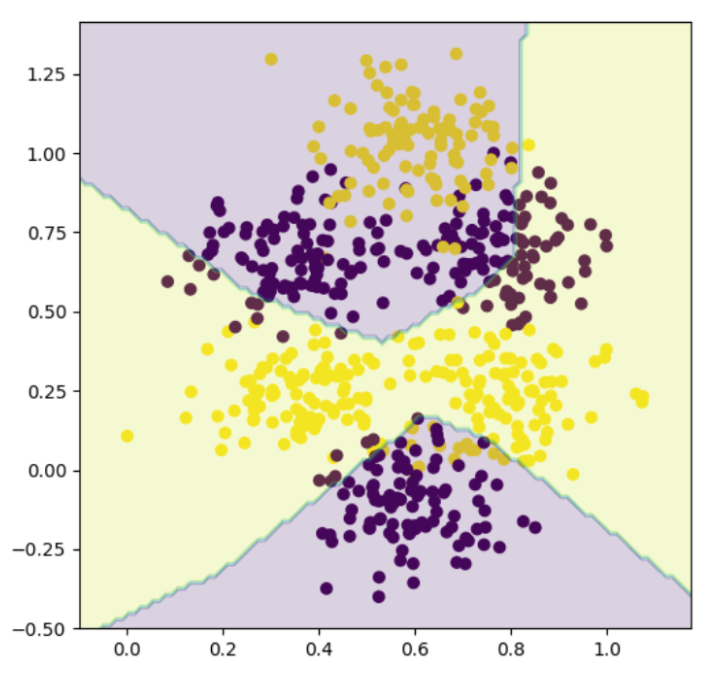}
    \includegraphics[width=2.7cm]{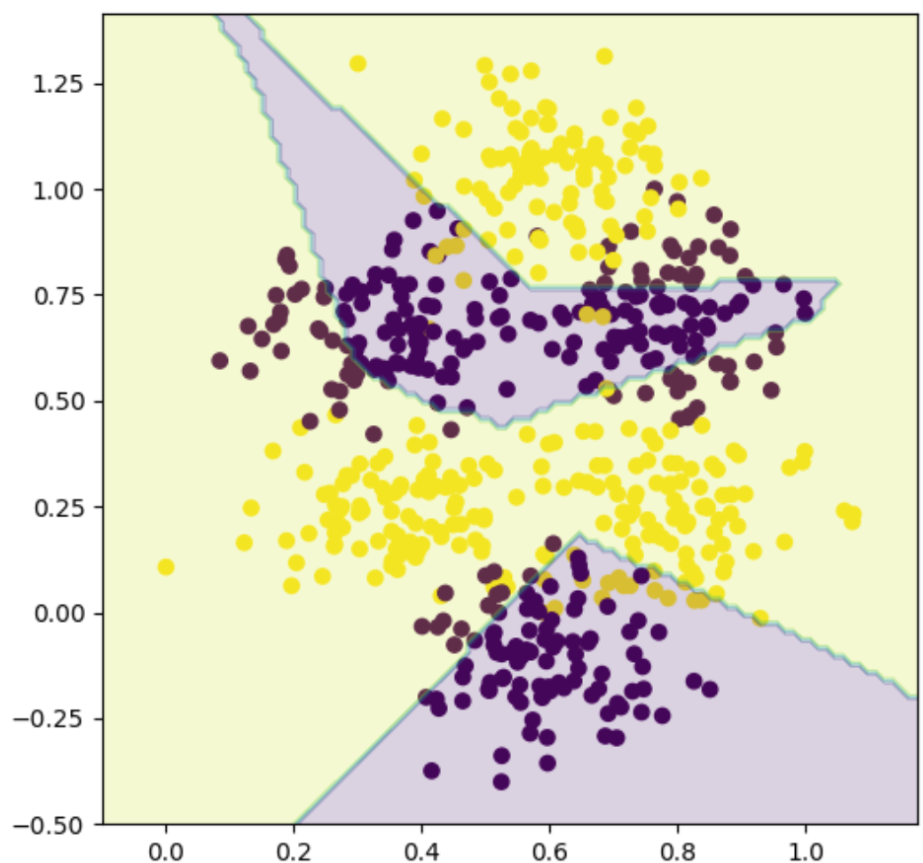}
    \includegraphics[width=2.8cm]{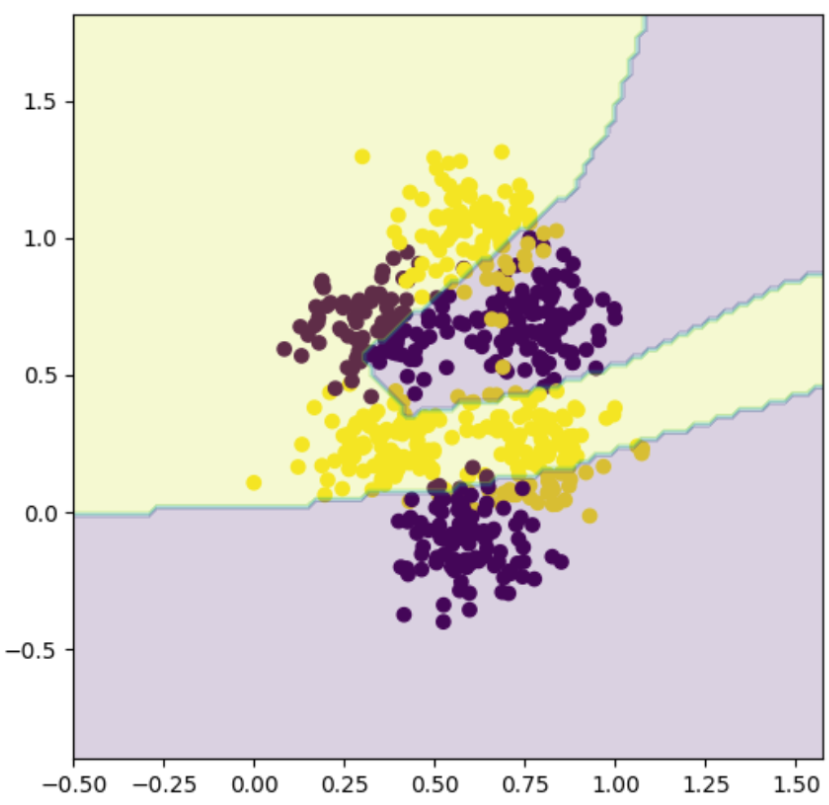}
    \caption{First row: Three different clients with totally different data distributions. Second row (from left): FedAvg after 40 communication steps, FedAvg after 100 communication steps, and FedGenP after 40 communication steps. }
    \label{fig: toy}
\end{figure}

We also perform experiments using synthetic data to observe the output of the proposed method and to what degree it matches our initial expectations. Illustrated in Figure \ref{fig: toy}, we examine a scenario involving three clients, each with two classes. Although the number of samples per class remains consistent, the distribution of these samples varies significantly across the clients.

The first row shows the data from each of the three clients. The second row of Figure \ref{fig: toy} shows the decision boundaries of the global models learned through the proposed method (on the right) and FedAvg (left and middle). It is evident that the proposed method achieves a satisfactory decision boundary with just 40 communication steps, whereas FedAvg requires 100 rounds to reach a comparable result.

\section{Discussion and Future Work}
In the proposed method, the server has to train a generator per client. In the case of a Cross-silo setting, in which clients are, for instance, companies, and the number of clients is not very large, this approach is feasible. In cross-device settings, with a large number of small clients, however,  this approach requires reconsideration due to the significant burden it places on the server. Therefore, in future research, we plan to enhance our method by extending it to clustered FL.

In clustered FL, clients are organized into groups based on their similarities. Subsequently, a global model is trained for each cluster. Typically, clients with similar characteristics are grouped together to mitigate the potential adverse effects of heterogeneity among clients. The literature suggests various criteria for determining these similarities.

Regardless of the similarity metric chosen, grouping clients in clustered FL can restrict the flow of information between certain clients, leading to potential information loss. To address this limitation and extend the scope of this paper, the proposed future approach involves constructing a generator for each cluster; alternatively, a conditional generator conditioned on different clusters. These generators are designed to incorporate information from other clusters into each specific cluster, mitigating the risk of information loss.

To achieve this, the design of the similarity metric should align with the generator. Consequently, the clustering similarity metric, when employing FedGenP, can focus on the similarity between gradient directions, given the significance of gradient directions for the generator. The clients with the same gradient direction should be grouped.

\section{Conclusion}
In this paper, we consider heterogeneous federated learning, where heterogeneity is not only due to imbalanced classes but also to the distribution of samples over the input space. For this reason, we examined the problem from the domain adaptation perspective and focused on minimizing the conflict between the clients' models. We theoretically show that the performance of the global model on every client is upper bounded by the heterogeneity between that client and the rest. To eliminate heterogeneity, we proposed personalized generative learning in which the server constructs a generator for each client. As a result of this generator, data is generated in the part of the space where the client's model conflicts with the rest. Thus, combining the original data with generated data will facilitate the clients' adaptation to global data, thereby enhancing the performance of the global model.

\section{Acknowledgments}
The work was carried out with support from The Knowledge Foundation and from Vinnova (Sweden's innovation agency) through the Vehicle Strategic Research and Innovation programme, FFI.

\bibliography{aaai24}

\begin{thebibliography}{30}
\providecommand{\natexlab}[1]{#1}

\bibitem[{Ben-David et~al.(2006)Ben-David, Blitzer, Crammer, and
  Pereira}]{ben2006analysis}
Ben-David, S.; Blitzer, J.; Crammer, K.; and Pereira, F. 2006.
\newblock Analysis of representations for domain adaptation.
\newblock \emph{Advances in neural information processing systems}, 19.

\bibitem[{Blitzer et~al.(2007)Blitzer, Crammer, Kulesza, Pereira, and
  Wortman}]{blitzer2007learning}
Blitzer, J.; Crammer, K.; Kulesza, A.; Pereira, F.; and Wortman, J. 2007.
\newblock Learning bounds for domain adaptation.
\newblock \emph{Advances in neural information processing systems}, 20.

\bibitem[{Caldas et~al.(2018)Caldas, Duddu, Wu, Li, Kone{\v{c}}n{\`y}, McMahan,
  Smith, and Talwalkar}]{caldas2018leaf}
Caldas, S.; Duddu, S. M.~K.; Wu, P.; Li, T.; Kone{\v{c}}n{\`y}, J.; McMahan,
  H.~B.; Smith, V.; and Talwalkar, A. 2018.
\newblock Leaf: A benchmark for federated settings.
\newblock \emph{arXiv preprint arXiv:1812.01097}.

\bibitem[{Chen et~al.(2023)Chen, Frikha, Krompass, Gu, and
  Tresp}]{chen2023fraug}
Chen, H.; Frikha, A.; Krompass, D.; Gu, J.; and Tresp, V. 2023.
\newblock Fraug: Tackling federated learning with non-iid features via
  representation augmentation.
\newblock In \emph{Proceedings of the IEEE/CVF International Conference on
  Computer Vision}, 4849--4859.

\bibitem[{Cohen et~al.(2017)Cohen, Afshar, Tapson, and
  Van~Schaik}]{cohen2017emnist}
Cohen, G.; Afshar, S.; Tapson, J.; and Van~Schaik, A. 2017.
\newblock EMNIST: Extending MNIST to handwritten letters.
\newblock In \emph{2017 international joint conference on neural networks
  (IJCNN)}, 2921--2926. IEEE.

\bibitem[{de~Luca et~al.(2022)de~Luca, Zhang, Chen, and Yu}]{de2022mitigating}
de~Luca, A.~B.; Zhang, G.; Chen, X.; and Yu, Y. 2022.
\newblock Mitigating data heterogeneity in federated learning with data
  augmentation.
\newblock \emph{arXiv preprint arXiv:2206.09979}.

\bibitem[{Duan et~al.(2020)Duan, Liu, Chen, Liu, Tan, and Liang}]{duan2020self}
Duan, M.; Liu, D.; Chen, X.; Liu, R.; Tan, Y.; and Liang, L. 2020.
\newblock Self-balancing federated learning with global imbalanced data in
  mobile systems.
\newblock \emph{IEEE Transactions on Parallel and Distributed Systems}, 32(1):
  59--71.

\bibitem[{Ganin et~al.(2016)Ganin, Ustinova, Ajakan, Germain, Larochelle,
  Laviolette, Marchand, and Lempitsky}]{ganin2016domain}
Ganin, Y.; Ustinova, E.; Ajakan, H.; Germain, P.; Larochelle, H.; Laviolette,
  F.; Marchand, M.; and Lempitsky, V. 2016.
\newblock Domain-adversarial training of neural networks.
\newblock \emph{The journal of machine learning research}, 17(1): 2096--2030.

\bibitem[{Gao, Yao, and Yang(2022)}]{gao2022survey}
Gao, D.; Yao, X.; and Yang, Q. 2022.
\newblock A Survey on Heterogeneous Federated Learning.
\newblock \emph{arXiv preprint arXiv:2210.04505}.

\bibitem[{Gong et~al.(2022)Gong, Sharma, Karanam, Wu, Chen, Doermann, and
  Innanje}]{gong2022preserving}
Gong, X.; Sharma, A.; Karanam, S.; Wu, Z.; Chen, T.; Doermann, D.; and Innanje,
  A. 2022.
\newblock Preserving privacy in federated learning with ensemble cross-domain
  knowledge distillation.
\newblock In \emph{Proceedings of the AAAI Conference on Artificial
  Intelligence}, volume~36, 11891--11899.

\bibitem[{Hsu, Qi, and Brown(2019)}]{hsu2019measuring}
Hsu, T.-M.~H.; Qi, H.; and Brown, M. 2019.
\newblock Measuring the effects of non-identical data distribution for
  federated visual classification.
\newblock \emph{arXiv preprint arXiv:1909.06335}.

\bibitem[{Huang, Ye, and Du(2022)}]{huang2022learn}
Huang, W.; Ye, M.; and Du, B. 2022.
\newblock Learn from others and be yourself in heterogeneous federated
  learning.
\newblock In \emph{Proceedings of the IEEE/CVF Conference on Computer Vision
  and Pattern Recognition}, 10143--10153.

\bibitem[{Jeong et~al.(2018)Jeong, Oh, Kim, Park, Bennis, and
  Kim}]{jeong2018communication}
Jeong, E.; Oh, S.; Kim, H.; Park, J.; Bennis, M.; and Kim, S.-L. 2018.
\newblock Communication-efficient on-device machine learning: Federated
  distillation and augmentation under non-iid private data.
\newblock \emph{arXiv preprint arXiv:1811.11479}.

\bibitem[{LeCun et~al.(2010)LeCun, Cortes, Burges et~al.}]{lecun2010mnist}
LeCun, Y.; Cortes, C.; Burges, C.; et~al. 2010.
\newblock MNIST handwritten digit database.

\bibitem[{Li, Chen, and Lu(2023)}]{li2023feddkd}
Li, X.; Chen, B.; and Lu, W. 2023.
\newblock FedDKD: Federated learning with decentralized knowledge distillation.
\newblock \emph{Applied Intelligence}, 1--17.

\bibitem[{Lin et~al.(2020)Lin, Kong, Stich, and Jaggi}]{lin2020ensemble}
Lin, T.; Kong, L.; Stich, S.~U.; and Jaggi, M. 2020.
\newblock Ensemble distillation for robust model fusion in federated learning.
\newblock \emph{Advances in Neural Information Processing Systems}, 33:
  2351--2363.

\bibitem[{Ma et~al.(2022)Ma, Zhu, Lin, Chen, and Qin}]{ma2022state}
Ma, X.; Zhu, J.; Lin, Z.; Chen, S.; and Qin, Y. 2022.
\newblock A state-of-the-art survey on solving non-IID data in Federated
  Learning.
\newblock \emph{Future Generation Computer Systems}, 135: 244--258.

\bibitem[{McMahan et~al.(2017)McMahan, Moore, Ramage, Hampson, and
  y~Arcas}]{mcmahan2017communication}
McMahan, B.; Moore, E.; Ramage, D.; Hampson, S.; and y~Arcas, B.~A. 2017.
\newblock Communication-efficient learning of deep networks from decentralized
  data.
\newblock In \emph{Artificial intelligence and statistics}, 1273--1282. PMLR.

\bibitem[{Taghiyarrenani(2022)}]{taghiyarrenani2022learning}
Taghiyarrenani, Z. 2022.
\newblock \emph{Learning from Multiple Domains}.
\newblock Ph.D. thesis, Halmstad University Press.

\bibitem[{Taghiyarrenani and Berenji(2022)}]{taghiyarrenani2022noise}
Taghiyarrenani, Z.; and Berenji, A. 2022.
\newblock Noise-robust representation for fault identification with limited
  data via data augmentation.
\newblock In \emph{PHM Society European Conference}, volume~7, 473--479.

\bibitem[{Taghiyarrenani, Nowaczyk, and
  Pashami(2023)}]{taghiyarrenani2023analysis}
Taghiyarrenani, Z.; Nowaczyk, S.; and Pashami, S. 2023.
\newblock Analysis of Statistical Data Heterogeneity in Federated Fault
  Identification.
\newblock In \emph{PHM Society Asia-Pacific Conference}, volume~4.

\bibitem[{Tan et~al.(2022)Tan, Yu, Cui, and Yang}]{tan2022towards}
Tan, A.~Z.; Yu, H.; Cui, L.; and Yang, Q. 2022.
\newblock Towards personalized federated learning.
\newblock \emph{IEEE Transactions on Neural Networks and Learning Systems}.

\bibitem[{Wen et~al.(2022)Wen, Wu, Li, and Duan}]{wen2022communication}
Wen, H.; Wu, Y.; Li, J.; and Duan, H. 2022.
\newblock Communication-efficient federated data augmentation on non-iid data.
\newblock In \emph{Proceedings of the IEEE/CVF Conference on Computer Vision
  and Pattern Recognition}, 3377--3386.

\bibitem[{Wu et~al.(2020)Wu, Chen, Zhou, and Zhang}]{wu2020fedhome}
Wu, Q.; Chen, X.; Zhou, Z.; and Zhang, J. 2020.
\newblock Fedhome: Cloud-edge based personalized federated learning for in-home
  health monitoring.
\newblock \emph{IEEE Transactions on Mobile Computing}, 21(8): 2818--2832.

\bibitem[{Yao et~al.(2021)Yao, Pan, Dai, Wan, Ding, Jin, Xu, and
  Sun}]{yao2021local}
Yao, D.; Pan, W.; Dai, Y.; Wan, Y.; Ding, X.; Jin, H.; Xu, Z.; and Sun, L.
  2021.
\newblock Local-global knowledge distillation in heterogeneous federated
  learning with non-iid data.
\newblock \emph{arXiv preprint arXiv:2107.00051}.

\bibitem[{Ye et~al.(2023)Ye, Fang, Du, Yuen, and Tao}]{ye2023heterogeneous}
Ye, M.; Fang, X.; Du, B.; Yuen, P.~C.; and Tao, D. 2023.
\newblock Heterogeneous federated learning: State-of-the-art and research
  challenges.
\newblock \emph{ACM Computing Surveys}, 56(3): 1--44.

\bibitem[{Yoon et~al.(2021)Yoon, Shin, Hwang, and Yang}]{yoon2021fedmix}
Yoon, T.; Shin, S.; Hwang, S.~J.; and Yang, E. 2021.
\newblock Fedmix: Approximation of mixup under mean augmented federated
  learning.
\newblock \emph{arXiv preprint arXiv:2107.00233}.

\bibitem[{Zhang et~al.(2017)Zhang, Cisse, Dauphin, and
  Lopez-Paz}]{zhang2017mixup}
Zhang, H.; Cisse, M.; Dauphin, Y.~N.; and Lopez-Paz, D. 2017.
\newblock mixup: Beyond empirical risk minimization.
\newblock \emph{arXiv preprint arXiv:1710.09412}.

\bibitem[{Zhao et~al.(2018)Zhao, Li, Lai, Suda, Civin, and
  Chandra}]{zhao2018federated}
Zhao, Y.; Li, M.; Lai, L.; Suda, N.; Civin, D.; and Chandra, V. 2018.
\newblock Federated learning with non-iid data.
\newblock \emph{arXiv preprint arXiv:1806.00582}.

\bibitem[{Zhu, Hong, and Zhou(2021)}]{zhu2021data}
Zhu, Z.; Hong, J.; and Zhou, J. 2021.
\newblock Data-free knowledge distillation for heterogeneous federated
  learning.
\newblock In \emph{International Conference on Machine Learning}, 12878--12889.
  PMLR.

\end{thebibliography}

\end{document}